\documentclass[10pt,twocolumn,letterpaper]{article}

\usepackage[pagenumbers]{cvpr} 

\makeatletter
\@namedef{ver@everyshi.sty}{}
\makeatother
\usepackage{tikz}

\usepackage[accsupp]{axessibility}
\usepackage{graphicx}
\usepackage{amsmath}
\usepackage{amssymb}
\usepackage{booktabs}
\usepackage{graphicx}

\usepackage{comment}
\usepackage{amsmath} 
\usepackage{xspace}
\usepackage{color}
\usepackage{multirow}
\usepackage{subcaption}
\usepackage{multicol}
\usepackage{placeins}
\usepackage{float}
\usepackage[pagebackref,breaklinks,colorlinks]{hyperref}

\usepackage[capitalize]{cleveref}
\crefname{section}{Sec.}{Secs.}
\Crefname{section}{Section}{Sections}
\Crefname{table}{Table}{Tables}
\crefname{table}{Tab.}{Tabs.}

\def\cvprPaperID{9002} 
\def\confName{CVPR}
\def\confYear{2023}
\usepackage{enumitem}
\usepackage[flushleft]{threeparttable}

\makeatletter
\DeclareRobustCommand\onedot{\futurelet\@let@token\@onedot}
\def\@onedot{\ifx\@let@token.\else.\null\fi\xspace}

\def\eg{\emph{e.g}\onedot} 
\def\ie{\emph{i.e}\onedot} 
 
\def\etc{\emph{etc}\onedot} 
 
\def\etal{\emph{et al}\onedot}
\makeatother
\newcommand{\lb}{\vspace*{0.4em}\\}

\begin{document}

\def\ECCVSubNumber{4178}  

\title{WildLight: In-the-wild Inverse Rendering with a Flashlight} 

\author{Ziang Cheng, Junxuan Li, Hongdong Li\\
Australian National University\\
{\tt\small \{ziang.cheng,junxuan.li,hongdong.li\}@anu.edu.au}
}

\twocolumn[{%
\renewcommand\twocolumn[1][]{#1}%
\maketitle
\begin{center}
    \centering
    \captionsetup{type=figure}
    \includegraphics[width=\textwidth]{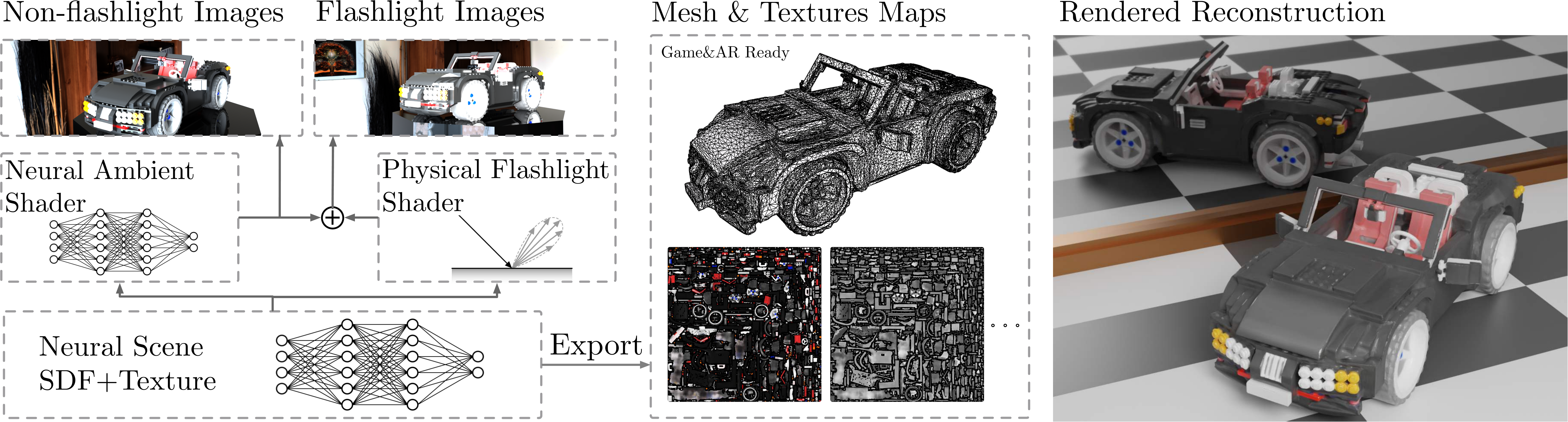}
    \captionof{figure}{Our method reconstructs object geometry and reflectance from unstructured flashlight and non-flashlight images captured under unknown environment lights. We use a co-located camera-light configuration that is available with most smartphones. Reconstructed objects can be easily converted to industry-ready triangle meshes and PBR textures. The image to the right is rendered from our mesh export.}
\end{center}%
}]

\begin{abstract}
This paper proposes a practical photometric solution for the challenging problem of in-the-wild inverse rendering under unknown ambient lighting. Our system recovers scene geometry and reflectance using only multi-view images captured by a smartphone. The key idea is to exploit smartphone's built-in flashlight as a minimally controlled light source, and decompose image intensities into two photometric components -- a static appearance corresponds to ambient flux, plus a dynamic reflection induced by the moving flashlight.  Our method does not require flash/non-flash images to be captured in pairs. Building on the success of neural light fields, we use an off-the-shelf method to capture the ambient reflections, while the flashlight component enables physically accurate photometric constraints to decouple reflectance and illumination. Compared to existing inverse rendering methods, our setup is applicable to non-darkroom environments yet sidesteps the inherent difficulties of explicit solving ambient reflections. We demonstrate by extensive experiments that our method is easy to implement, casual to set up, and consistently outperforms existing in-the-wild inverse rendering techniques. Finally, our neural reconstruction can be easily exported to PBR textured triangle mesh ready for industrial renderers. Our source codes a data are released to \url{https://github.com/za-cheng/WildLight}.
\end{abstract}

\section{Introduction}
{\em Rendering}  in computer graphics refers to computer generating photo-realistic images from known properties of a scene including scene geometry, materials, lighting, as well as camera parameters.  In contrast, inverse rendering is regarded as a computer vision task, whose aim is to recover these unknown properties of a scene from images. Due to the ill-posed nature of inverse rendering, most existing methods concede the fullest solution and instead tackle only a simplified, partial problem; for instance, assuming simplified, known lighting conditions, initial geometry, or with diffuse or low specular materials.

Traditional photometric methods for solving inverse rendering (such as photometric stereo) are often restricted to laboratory settings: image intensities are measured by high dynamic range cameras, under controlled illumination conditions and without ambient light contamination. Moreover, many methods rely on the availability of a good initial estimation to start the optimization process (\eg~\cite{nam2018practical,li2020multi}) or assume purely diffuse (Lambertian) reflections (\eg~\cite{park2016robust,logothetis2019differential}). While recent neural-net methods are able to handle specular highlights and complex geometries, \eg~\cite{bi2020neural,cheng2021one,zhang2022iron,li2018learning}), they are still restricted to a laboratory darkroom environment, hindering their practicality.

Conversely, neural rendering and appearance learning techniques have the ability to work outside darkroom. This is achieved by bypassing the physical reflection model. They instead learn the illumination-specific appearance (\ie light field) of the scene conditioned on a geometric representation (\eg~\cite{mildenhall2020nerf,yariv2020multiview,niemeyer2020differentiable}). While these representations are empirically powerful in recovering complex scene geometry and appearance, they cannot separate reflective properties from illumination. Furthermore, the underlying geometry is provably ill-constrained. 
Attempts have been made to physically decompose the appearance into reflectance and environment illumination to support in-the-wild inverse rendering \cite{zhang2021physg,boss2021nerd,zhang2021nerfactor,srinivasan2021nerv,munkberg2021nvdiffrec}. However, this presents a much more complex and similarly ill-posed problem due to unknown ambient illumination. Consequently, these methods still trail behind traditional photometric methods in terms of accuracy and robustness.

This paper aims to fill the gap between conventional darkroom methods and in-the-wild inverse rending, and offer a solution that combines the best of both worlds, \ie being practical, well-posed and easy-to-solve at the same time. Instead of attempting to directly decouple reflectance from the unknown ambient illumination, we learn the ambient reflection with a neural light field, and exploit an additional, minimally controlled light source, being the smartphone's flashlight, for physical constraints on reflectance. During the image capture stage, we take some images with the flashlight turned on, and others with the flashlight off, all at free viewpoints. Images without the flashlight describe the ambient reflections only, while the images with flashlight are the photometric summation of both ambient and flashlight reflections. We learn the ambient component with an off-the-shelf neural novel-view-synthesis technique \cite{wang2021neus}, and delegate the flashlight component to a physically-based reflection model for inferring reflectance. Both ambient and flashlight components are conditioned on a unified scene intrinsic network that predicts scene geometry and reflectance distribution. Our method is easy to set up, easy to implement, and consistently outperforms competing state-of-the-arts. The reconstructed objects can be directly plugged into game/rendering engines as high fidelity virtual assets in the standard format of textured meshes.

\section{Related work}
Inverse rendering is a long-standing and highly researched topic. To limit the scope of discussion, below we only give a partial review of RGB camera-based multi-view 3D reconstruction methods.

\paragraph{Darkroom methods.} Traditional photometric methods assume a darkroom environment where the lighting condition is controlled and calibrated. Earlier methods assume a mostly diffuse appearance to simplify the reflection model  \cite{higo2009hand,vlasic2009dynamic,wu2010fusing,delaunoy2014photometric,park2016robust,logothetis2019differential,oxholm2012shape}. Recent papers are able to deal with moderately glossy objects by incorporating explicit specular reflectance, typically controlled by a roughness parameter. However, the underlying problem becomes highly non-convex under this setting, and many methods require an initial geometry to bootstrap the optimization process. Initial geometry is acquired either from multi-view stereo/structure-from-motion \cite{nam2018practical,li2020multi,Zhou2013multi}, or from RGB-D sensors \cite{maier2017intrinsic3d,bylow2019combining,schmitt2020joint,ha2020progressive}. Notably, many recent methods are based on a co-located camera/light scanner to simplify imaging setup and reflection model. Nam \etal~\cite{nam2018practical} propose a optimization pipeline that iteratively refines initial geometry and reflectance under co-located reflection. \cite{bi2020deep,bi2020deep2,bi2020neural} use deep neural networks to learn representations of geometry and reflectance, supervised by ground truth, or directly by multi-view images. Cheng~\etal~\cite{cheng2021multi} proposed an initialization-free optimization framework for highly specular surfaces, and recently developed a neural spherical parameterization method for learning shape and reflectance \cite{cheng2021one}. Luan~\etal~\cite{luan2021unified} used an Monte Carlo edge sampling approach for joint optimization of shape and reflectance. Recently, Zhang~\etal~\cite{zhang2022iron} recovers objects as neural SDF and materials, and use an edge-aware renderer for refinement. All above methods are however restricted to a darkroom environment, and require per-view object masks to bootstrap geometry optimization.

\paragraph{Neural appearance/light field.} Under ambient illumination, scene radiance distribution features strong co-relation with geometry. Neural implicit methods exploit this prior to model appearance as a light field conditioned on geometry. DVR \cite{niemeyer2020differentiable} proposed a differentiable renderer for implicit occupancy networks. NeRF \cite{mildenhall2020nerf} use an MLP to represent density and view-dependent radiance fields supervised by input images. However, the underlying geometry is provably under-constrained and often of low quality. IDR \cite{yariv2020multiview} alleviates this problem by employing an signed distance field (SDF) representation that defines hard geometrical surfaces, and use a neural network to learn reflected lights. NeuS \cite{wang2021neus} proposes an alternative rendering technique for SDF based on volumetric rendering. In a similar spirit, UniSurf \cite{oechsle2021unisurf} unifies surface and volumetric rendering for an occupancy network representation. Kaya \etal~\cite{kaya2022neural} extract surface normal from photometric stereo, on which a NeRF-based neural shader is trained. While empirically these methods achieved high quality results for geometry and appearance reconstructions, they do not follow a physically based reflection model and cannot separate reflectance from illumination conditions to support relighting. 

\paragraph{Inverse rendering under ambient light.} Unlike the darkroom setup, in-the-wild inverse rendering is an ill-posed and much more challenging problem. Part of the challenge arises from the fact that reflected lights need to be integrated over all incident directions, which is a highly expensive operation. As a result, the ambient lighting is approximated by low resolution or low frequency environment maps, and all light sources are assumed to be infinitely far away. Early work of Zhang~\etal~\cite{zhang2003shape} is restricted to Lambertian objects and small camera motions. Oxholm~\etal~\cite{oxholm2014multiview} proposed an energy based approach to solve for simple geometries initialized from visual hull. NeRFactor \cite{zhang2021nerfactor} first extracts geometry from pretrained NeRF then decomposes radiance into visibility, reflectance and illumination for joint optimization. NeRV \cite{srinivasan2021nerv} learns a neural visibility field for efficiently computing secondary reflections. PhySG \cite{zhang2021physg} sidesteps the numerical integration using Spherical Gaussian BRDF and illumination maps. NerD \cite{boss2021nerd} adds support for varying illuminations, and NeuralPIL \cite{boss2021neuralpil} uses an MLP to learn the light integration. Recently, Munkberg~\etal~\cite{munkberg2021nvdiffrec} combine neural representation with mesh-based rasterizer for differentiable rendering.  While these methods are flexible in imaging setup, their accuracy trails behind conventional darkroom based approaches due to the intrinsic difficulty of factorizing ambient reflections. Furthermore, NeRF-based methods \cite{boss2021nerd,zhang2021nerfactor,srinivasan2021nerv,boss2021neuralpil} cannot uphold a hard, smooth surface boundary.

\paragraph{Our approach.} We adopt a hybrid approach that combines traditional darkroom photometric methods and neural light fields. Compared to existing  in-the-wild inverse rendering methods, our approach requires an additional light source, but is not restricted to the distant ambient lighting assumption, can work without object masks, and avoids the difficulties of factorizing ambient reflections. On the other hand, the flashlight still offers sufficient photometric constraint to solve for reflectance and regulate geometry, achieving comparable accuracy to darkroom-based methods while being arguably more flexible. 

\section{Photometric image formulation}
Our method approximates the scene as surfaces of opaque appearance. The scene is under a static but unknown ambient illumination (\ie, the environment map), and may be further illuminated by a moving point light source that is co-located with the camera. With such assumptions in mind, scene radiance captured by the camera is the summation of two photometric components: an ambient component corresponding to flux transmitted from all light emitters within the scene, and an optional flashlight component that accounts for flashlight energy reflected off the surface. 

More formally, with a smartphone device, we assume its flashlight is a point light source co-located with the camera. The raw image intensity of surface point $\mathbf{x}\in\mathbb{R}^3$ at view direction $\mathbf{v}$ and viewing distance $t$ is expressed as the summation of an ambient term $\mathcal{A}$ and a flashlight term $\mathcal{L}$:
\begin{equation}
    \mathcal{I}(\mathbf{x}, \mathbf{v}, t, s) = \mathcal{A}(\mathbf{x}, \mathbf{v}) + s\gamma\mathcal{L}(\mathbf{x}, \mathbf{v}, t), \label{eq:formation}
\end{equation}
where $s\in\{0,1\}$ is a binary switch indicating whether the flashlight is on, and $\gamma$ is the unknown flashlight intensity. Ambient term defines a scene appearance independent of the flashlight constituting non-flash images. Neural appearance/novel-view synthesis methods learn $\mathcal{A}$ as a black box function to supervise an underlying scene geometry that $\mathcal{A}$ is conditioned on (\ie spatial distribution of $\mathbf{x}$) \cite{mildenhall2020nerf,yariv2020multiview}. This problem however, is well known to be under-constrained due to the view dependency of $\mathcal{A}$, in which case the geometry cannot be well recovered \cite{zhang2020nerf++}. An alternative approach is to explain $\mathcal{A}$ with an ambient reflection model to obtain physical constrains, but this involves a much more complicated problem that is also often ill-posed \cite{oxholm2014multiview,srinivasan2021nerv}. 

To mitigate above issues, we incorporate an additional flashlight reflection term. It follows real world physics but is much simpler and better posed than ambient reflections
\begin{equation}
  \mathcal{L}(\mathbf{x}, \mathbf{v}, t) = \frac{\mathcal{E_\mathbf{v}}}{t^2} \rho_\mathbf{x}(\mathbf{n_x}, \mathbf{v},\mathbf{v})\max(\mathbf{n}_\mathbf{x}^\top\mathbf{v},0). \label{eq:reflection}
\end{equation}
$\mathcal{E_\mathbf{v}}$ denotes the flashlight radiant intensity at direction $\mathbf{v}$, and $\rho_\mathbf{x}$ is the material's reflectance function of three directions: surface normal vector $\mathbf{n_x}$, and view and light directions joint at $\mathbf{v}.$\footnote{Actual difference between and view and light angles is around 1 degree for an iPhone at $0.5m$ view distance.} . The reflectance function can be further parameterized as
\begin{equation}
\rho_\mathbf{x}(\mathbf{n},\mathbf{v},\mathbf{l})=\rho(\mathbf{n},\mathbf{v},\mathbf{l};\Theta_\mathbf{x})
\end{equation}
such that it is defined by a set of parameters $\Theta_\mathbf{x}$. Without loss of generality, we further assume the flashlight exhibits an isotropic distribution of intensity within the field of view.\footnote{Any radial anisotopicity can be corrected by applying per-pixel anti-vignetting compensation factors, as long as light and camera are co-located.} That is, we may simply define $\mathcal{E_\mathbf{v}}=1$ as one unit of radiant intensity.

The co-located reflection model in (\ref{eq:reflection}) provides strong photometric constraints on the scene geometry and reflection, allowing them to be recovered jointly if flashlight reflection $\mathcal{L}$ is made known \cite{nam2018practical,cheng2021multi,luan2021unified}. In reality, however, the flashlight images contain both flashlight reflections $\mathcal{L}$ and ambient reflections/emissions $\mathcal{A}$, and it is non-trivial to separate one from another. Many photometric methods are therefore limited to darkroom environment where there is no ambient light contamination. In this paper, we sidestep this problem by employing a neural appearance model to learn the ambient component.

\section{Scene representation and methodology}
We parameterize target object's geometry and reflectance within an \textit{intrinsic network}, and use an \textit{ambient network} to shade its ambient reflections~$\mathcal{A}$ as a neural light field. Our method supports both masked and mask-less object reconstruction. When object masks are not available, we only reconstruct the partial scene within a given spherical Region of Interest (ROI) recognized as foreground region. The flashlight and ambient lights outside this ROI are instead approximated by a \textit{background NeRF} inspired by Zhang \etal \cite{zhang2020nerf++}. With such foreground-background separation in mind, we rewrite (\ref{eq:formation}) as
\begin{equation}
    I_\mathbf{x} = \left\{\begin{array}{cl}
        \hat{\mathcal{A}} + s\gamma \hat{\mathcal{L}} & \text{if}\; \mathbf{x}\in \text{Foreground} \\
        \hat{A}_\text{NeRF} + s\gamma \hat{{L}}_\text{NeRF} & \text{if}\; \mathbf{x}\in \text{Background} 
    \end{array}\right.
\end{equation}

\subsection{Foreground model}
The foreground geometry and reflectance are represented by the intrinsic network, on which ambient and flashlight reflections are both conditioned.
\paragraph{\bf{Intrinsic network}} The \textit{intrinsic network} recovers object intrinsic properties (\ie~geometry and reflectance) independent of any illumination condition. The intrinsic network learns geometry as a signed distance field, and outputs reflectance parameters at given location. 
\begin{equation}
\hat{\mathcal{N}}: \{\mathbf{x}\}\rightarrow \{(distance,\Theta,\mathbf{f})\}.
\end{equation}
Apart from the signed distance and reflectance parameters $\Theta$, the network also outputs a feature vector $\mathbf{f}_\mathbf{x}$ as a general descriptor for the local scene around $\mathbf{x}$. Furthermore, the network derivative defines the normal direction of a surface point:
\begin{equation}
    \mathbf{n}_\mathbf{x} = \nabla_{\mathbf{x}} distance.
\end{equation}

\paragraph{\bf{Ambient appearance network}} The ambient network acts as a neural shader that learns ambient reflections as a view-dependent neural light field defined on scene geometry.
\begin{equation}
    \hat{\mathcal{A}}:  \{(\mathbf{x}, \mathbf{v}, \mathbf{n}_\mathbf{x}, \Theta_\mathbf{x}, \mathbf{f}_\mathbf{x})\}\rightarrow \{RGB\}.
\end{equation}
Compared to (\ref{eq:formation}), we incorporate the normal direction and reflectance parameters as additional inputs since ambient reflection is empirically co-related with them. This formulation is inherited from previous work in neural rendering \cite{yariv2020multiview,oechsle2021unisurf,wang2021neus}, except here the network also receives reflectance parameters $\Theta_\mathbf{x}$ as additional input. 

\paragraph{\bf{Flashlight reflection}}
While the ambient network regularizes scene geometry by the MLP's inherent prior, there exist many solutions of scene geometry and radiance field that lead to the same multi-view images. We overcome this ambiguity by lighting a subset of images with a flashlight. The flashlight reflection $\mathcal{L}$ provides important physical constraints that disambiguate geometry, and further disentangle reflectance from illumination.
\begin{equation}
    \hat{\mathcal{L}}(\mathbf{n}_\mathbf{x}, \mathbf{v}, t, \Theta_\mathbf{x}) = \frac{1}{t^2} \rho(\mathbf{n_x}, \mathbf{v},\mathbf{v};\Theta_\mathbf{x})\max(\mathbf{n}_\mathbf{x}^\top\mathbf{v},0)
\end{equation}
We use a physically based reflectance model to estimate the flashlight reflection. Specifically, we parameterize $\rho(\cdot)$ using Disney's principled BRDF model (also known as the PBR texture model) \cite{burley2012physically}, that is, a linear mix of two diffuse lobes and three specular lobes\footnote{We disable the transmission/refraction lobe in the original Disney's model since we only consider opaque objects.}:
\begin{itemize}[leftmargin=*]
    \item Unlike the simpler Lambertian model, the diffuse lobe accounts for varying retro-reflections at grazing angles. The diffuse model also blends in a secondary subsurface lobe for modeling scattering effects observed in translucent materials (\eg~human skin).
    \item The main specular model follows the GGX microsurface distribution \cite{walter2007microfacet} and has two lobes: a tinted lobe for metals, and an achromatic lobe for dielectric materials.
    \item Another achromatic specular lobe is included for clearcoat materials. This lobe follows Berry's microsurface distribution, and has a roughness independent of other lobes.
\end{itemize}
Since the flashlight and camera are co-located, we slightly modify \cite{burley2012physically} to use a joint mask-shadowing term for all specular lobes and include the flashlight intensity $\gamma$ as a trainable parameter. The reader is referred to the Appendix for the exact formulation of BRDF $\rho(\cdot;\Theta_\mathbf{x})$.

\subsection{Background NeRF}  The scene outside the foreground ROI is instead learned by a modified NeRF model\cite{zhang2020nerf++,mildenhall2020nerf}. There are two NeRFs responsible for ambient and flashlight components respectively:
\begin{align}
    \hat{I}_\text{NeRF}(\mathbf{x}, \mathbf{v}, t, s) =  \hat{A}_\text{NeRF}(\mathbf{x}, \mathbf{v}) + s\gamma \hat{L}_\text{NeRF}(\mathbf{x},\mathbf{v}),t) \label{eq:nerf}\\
    \text{where}\;\hat{L}_\text{NeRF}(\mathbf{x},\mathbf{v}),t) = \frac{1}{t^2}\rho_\text{NeRF}(\mathbf{x},\mathbf{v})).
\end{align}
The network $\rho_\text{NeRF}$ is analogue to the reflectance function in (\ref{eq:formation}), except here the normal direction is not explicitly given but conditioned on the input $\mathbf{x}$. While
$\hat{A}_\text{NeRF}$ and $\rho_\text{NeRF}$ have different physical significance (radiance versus reflectance), implementation-wise we merge them into a single network with two RGB output branches: one for ambient radiance and one for reflectance, under the unified name NeRF (`R' stands for radiance or reflectance).
Unlike with the foreground model, the background NeRF does not necessarily define a geometrical surface, nor does it conform to a physically-based reflectance function. Instead, it is solely purposed to reproduce the background appearance from which the foreground can be easily separated.

\subsection{Training by rendering}
\paragraph{\bf{Volumetric renderer}}
\begin{figure}[!htb]
    \centering
    \includegraphics[trim={0.5cm 1.75cm 0.5cm 0}, clip, width=0.47\textwidth]{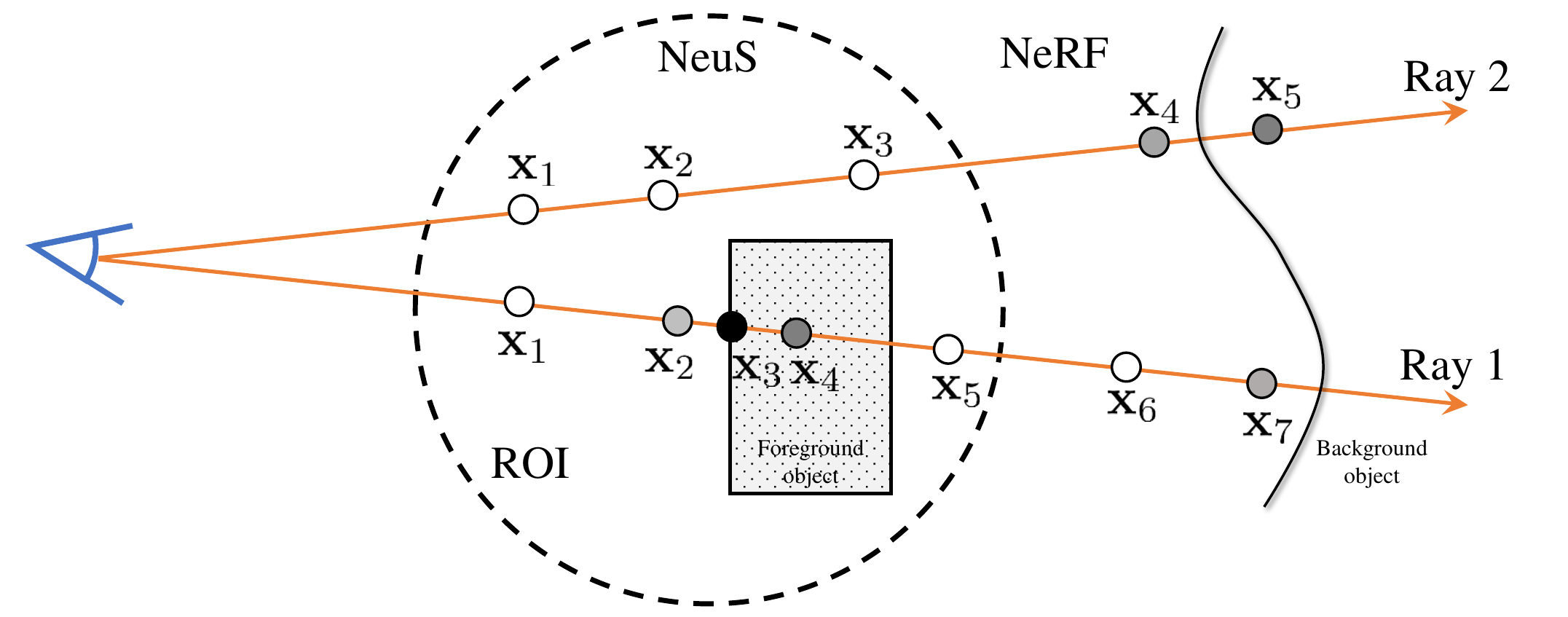}
    \caption{The hybrid rendering approach based on volumetric rendering: along each camera ray we densely trace multiple points $\mathbf{x}_1,...,\mathbf{x}_n$ in space, and render the RGB values as the weighted mean of per-point RGBs. Alpha distributions on the rays are colored in different shades (darker is greater). RGB values are aggregated using alpha composition, where RGB and alpha values are computed from NeuS \cite{wang2021neus} for points inside the pre-defined ROI, or from the background NeRF \cite{zhang2020nerf++} for points outside. }
    \label{fig:ray-tracer}
\end{figure}
We adopt an off-the-shelf SDF rendering technique called NeuS \cite{wang2021neus}. Compare to conventional SDF rendering algorithms that trace a single intersection point per viewing ray, NeuS samples multiple points per ray while only the points closer and more orthogonal to the surface are assigned greater weights. This enables gradients to be traced back not only on rays hitting the surface, but also on those tangent to self-occlusion boundaries (\eg for thin, concave geometries). \lb
Denote the camera center by $\mathbf{o}$, a viewing ray is parameterized as $\mathbf{x}(t)=\mathbf{o}-\mathbf{v}t$. We render its RGB intensity by sampling multiple points at $t_i\in[0,\infty)$ along the ray, and use the alpha-composition
\begin{align}
    \hat{\mathcal{I}}(\mathbf{o},\mathbf{v}, s) = \sum_i \omega_i\ \hat{\mathcal{I}}\big(\mathbf{x}(t_i), \mathbf{v}, t_i, s\big)\\\; \text{where} \; \omega_i=\alpha_i\prod_{t_j<t_i} (1-\alpha_j). \label{eq:alpha_composition}
\end{align}
When foreground masks are available, we turn off the background NeRF and render foreground rays only. Conversely, in the mask-less case, we use a hybrid rendering approach where $\alpha$ values and RGB intensities are obtained from NeuS \cite{wang2021neus} for points inside the foreground ROI, and from the background NeRF (\ref{eq:nerf}) for points outside. Figure \ref{fig:ray-tracer} illustrates the $\alpha$ distribution along viewing rays in the mask-less scenario.\\

\paragraph{\bf{Training loss}}
All networks are directly supervised from multi-view images, and the intrinsic network is regularized by an Eikonal loss and an optional foreground mask loss on the SDF. The overall training loss is
\begin{equation}
    Loss = Loss_\text{RGB} + w_E Loss_\text{Eikonal} + \underbrace{w_M Loss_\text{mask}}_{\text{optional}}
\end{equation}
where $Loss_\text{RGB}$ is a color saturation aware L$1$ loss for all pixels $p$.
\begin{equation}
    Loss_\text{RGB} = \frac{1}{|P|} \sum_{p\in P} M_p |\mathcal{I}_p-\hat{\mathcal{I}}_p|
\end{equation}
The binary variable $M_p$ is set to $1$ if and only if $\mathcal{I}_p$ or $\hat{\mathcal{I}}_p$ does not exceed the RGB color range of $[0,1)$. Otherwise $M_p=0$, so that the loss is turned off where both predicted and actual pixel values are saturated (\eg~on specular highlights). \\\\
The Eikonal loss modulates the gradients of SDF. To compute this loss, we uniformly sample points inside the ROI where the SDF is defined.
\begin{equation}
    Loss_\text{Eikonal}=\mathop{\mathbb{E}}_{\mathbf{x}\sim U}(1-\|\nabla_\mathbf{x}distance\|)^2
\end{equation}
Finally, when available, foreground masks can be used to supervise SDF to regulate geometry. The optional foreground mask loss is defined as:
\begin{equation}
    Loss_\text{mask} = \frac{1}{|P|} BCE(mask_p, \sum_i \omega^p_i),
\end{equation}
where $BCE$ is the binary cross entropy, and $\omega^p_i$ is defined as in (\ref{eq:alpha_composition}).

\paragraph{\bf{Mesh and texture extraction}} After training, scene geometry and surface reflectance can be extracted from the intrinsic network $\hat{\mathcal{N}}$ and stored into a standard mesh format supported by industrial renderers (\eg~ Blender, Unreal Engine \etc). We follow the procedure described below for mesh and PBR texture extraction:
\begin{enumerate}[leftmargin=*]
    \item A mesh is extracted from the zero isosurface of intrinsic SDF using marching cubes algorithm \cite{lewiner2003efficient}, followed by a mesh simplification step \cite{garland1997surface}. 
    \item Mesh vertices are assigned UV coordinates. A 3D atlas is interpolated from vertex UVs, that records for each pixel in UV space, its corresponding 3D location on the mesh surface.
    \item We feed the 3D atlas to the intrinsic network again to generate texture and normal maps.
\end{enumerate}
We note that many other methods lack the portability to external industrial renderers. Examples include NeRF-based methods \cite{boss2021nerd,boss2021neuralpil,zhang2021nerfactor,srinivasan2021nerv} based on the particle cloud model that is not commonly supported, and \cite{zhang2021physg} that uses a custom Spherical Gaussian reflectance model.
\section{Experiments}
\subsection{Training routine and hyper-parameters}
We use fixed weight $\omega_E=0.1$, and set non-zero $\omega_M=0.1$ only if masks are available. We train our networks by randomly sampling 256 rays per batch, where 128 points per ray are sampled within the ROI sphere. If the background NeRF is enabled, an extra 32 points per ray is sampled behind the ROI for training NeRF. Network weights are optimized with Adam \cite{kingma2014adam} at a learning rate of $5e-4$, which is gradually reduced to $2.5e-5$ after 5000 iterations. The networks are trained for a total 400,000 iterations (or 2 epochs) on synthetic scenes, and double that iterations on real scenes. Total training time varies between 12 to 24 hours on a single RTX 3090. The intrinsic network weights are initialized such that initial SDF roughly corresponds to the ROI sphere \cite{atzmon2020sal}.\lb
For each input image, we assume the camera parameters are known, and define the ROI sphere in world space such that the objects of interest are well contained in it.

\subsection{Dataset and image acquisition}
We evaluate our methods on both synthetic and real-world images in indoor settings. We will publish our datasets for reproducible research.
\begin{figure*}
    \centering
    \includegraphics[width=0.9\textwidth]{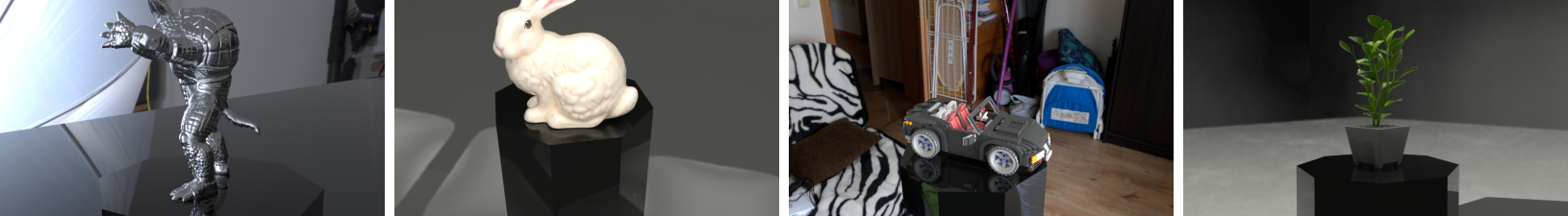}
    \caption{Example synthetic images under ambient lighting.}
    \label{fig:input_images}
\end{figure*}
\paragraph{Synthetic Image. } Our synthetic images are generated with Blender, using a co-located camera and flashlight configuration. We use an imaging setup consistent with most real world smartphones, where the virtual flashlight emits a cone of light that roughly covers the camera's field of view. The flashlight is set to operate at 0.2 to 0.8 watt, and objects are located at viewing distance of 0.3 to 0.5 meter. For the virtual environment setup, we put the objects on a virtual pedestal in indoor environments. For each object we generate 75 images under ambient lighting alone, and another 75 images with flashlight turned on. Viewpoints are randomly sampled from a half dome above the object with camera always facing object center. The synthetic dataset consists of a total of 150 HDR images at 1K resolution, the corresponding foreground masks, as well as ground truth camera parameters. Example images are shown in Figure \ref{fig:input_images}. Half of total images are lit by flashlight, and the other half by environment light only.
\paragraph{Real Image. } For real world image acquisition, we used an iPhone and the ``ProCam'' app to take images in RAW format with a linear camera response. During each capture, we maintained a fixed camera exposure time, focus, and white balance for all images. The camera response ratio was automatically adjusted per image and recorded in file metadata, which we later used to scale all images back to a unified intensity scale. The camera pose are acquired from an AR board behind the object. Images are taken inside an apartment unit under interior lighting. We place the camera at 0.2 to 0.5 meter viewing distance, and move the camera in a spiral pattern around the object. We take 56 to 112 HDR images per object. In lack of object segmentation masks, we enable the background NeRF model and turn off $Loss_\text{mask}$. Around half of total images are lit by flashlight, and the other half by environment light only.

\subsection{Evaluation and comparison}
To the best of our knowledge, there is no previous method specifically engineered for our imaging configuration (\ie in-the-wild images partly lit by flashlight). Therefore direct comparison with state-of-arts would not be fair or feasible. Instead, we keep all but a few imaging conditions identical, but allow following exceptions in favor of each competing method's own configuration:\\
\textbf{NeRD \cite{boss2021nerd}} assumes varying global illumination. Therefore we replace the flashlight with a moving distant light source whose direction is always aligned with camera's optical axis. We adjust the light power to a similar intensity to our flashlight. \\
\textbf{PhySG \cite{zhang2021physg}} assumes static global illumination. Therefore we remove the flashlight altogether and render images solely under ambient illumination.\\

\begin{table*}[!htb]
    \centering
    \caption{Quantitative comparison of geometry estimations on synthetic scenes. Distance errors are defined as the mesh-to-mesh distance on the visible parts of surface. \textbf{Distance and depth errors are measured in ratio to $10^{-3}$ of object lengths.} Normal errors are measured in degrees. Our method consistently outperforms both NeRD and PhySG by a substantial margin.}
    \label{tab:quantitative_comparison}
    \begin{threeparttable}
    \setlength\tabcolsep{0.4em}
\begin{tabular}{c*{9}{|c}}
\multirow{3}{*}{Objects} & \multicolumn{9}{c}{Mean/Median errors per method per metric} \\\cline{2-10}
& \multicolumn{3}{c|}{NeRD \cite{boss2021nerd}} & \multicolumn{3}{c|}{PhySG \cite{zhang2021physg}} & \multicolumn{3}{c}{Ours} \\\cline{2-10}
                  & {Distance*} & {Depth} & {Normal} & {Distance} & {Depth} & {Normal}  & {Distance} & {Depth} & {Normal} \\\cline{2-10}
                  \hline
Armadillo &  5.0 / 2.9   &  14.0 / 4.3 & 22.1 / 18.3   &  9.0 / 3.1  & 23.4 / 3.4 & 15.5 / 9.8 & \textbf{1.0} / \textbf{0.7} & \textbf{2.1} / \textbf{0.7} & \textbf{6.4} / \textbf{4.5}  \\       
Bunny & 64.9 / 48.7   &  168.2 / 79.4 & 69.8 / 68.5   &  3.8 / 1.7  & 7.4 / 1.7 & 7.7 / 5.0 & \textbf{1.5} / \textbf{1.0} & \textbf{2.6} / \textbf{1.4} & \textbf{4.5} / \textbf{3.3}  \\  
LegoCar & -   &  -   &  -   &  34.1 / 26.6 & 64.1 / 29.3 & 39.2 / 29.2 & \textbf{16.2} / \textbf{8.6} & \textbf{10.6} / \textbf{3.1} & \textbf{19.3} / \textbf{9.1}  \\  
Plant & 15.4 / 13.4   &  47.0 / 15.3 & 26.0 / 18.5   &  16.7 / 13.6  & 69.1 / 23.4 & 31.5 / 17.6 & \textbf{1.2} / \textbf{0.9} & \textbf{4.3} / \textbf{1.2} & \textbf{7.1} / \textbf{3.3}
\end{tabular}
\begin{tablenotes}
\small
    \item *NeRD does not define a surface geometry for direct comparison. To obtain a surface from NeRD, we extract a point cloud by tracing camera rays' expected depths, followed by outlier removal and Poisson surface reconstruction \cite{kazhdan2006poisson}.
    \item -NeRD failed on the LegoCar sequence and produced an empty scene with zero density.
  \end{tablenotes}
  \end{threeparttable}
\end{table*}
\begin{figure*}[!htb]
    \captionsetup[subfigure]{labelformat=empty}
    \centering
    \includegraphics[width=0.98\textwidth]{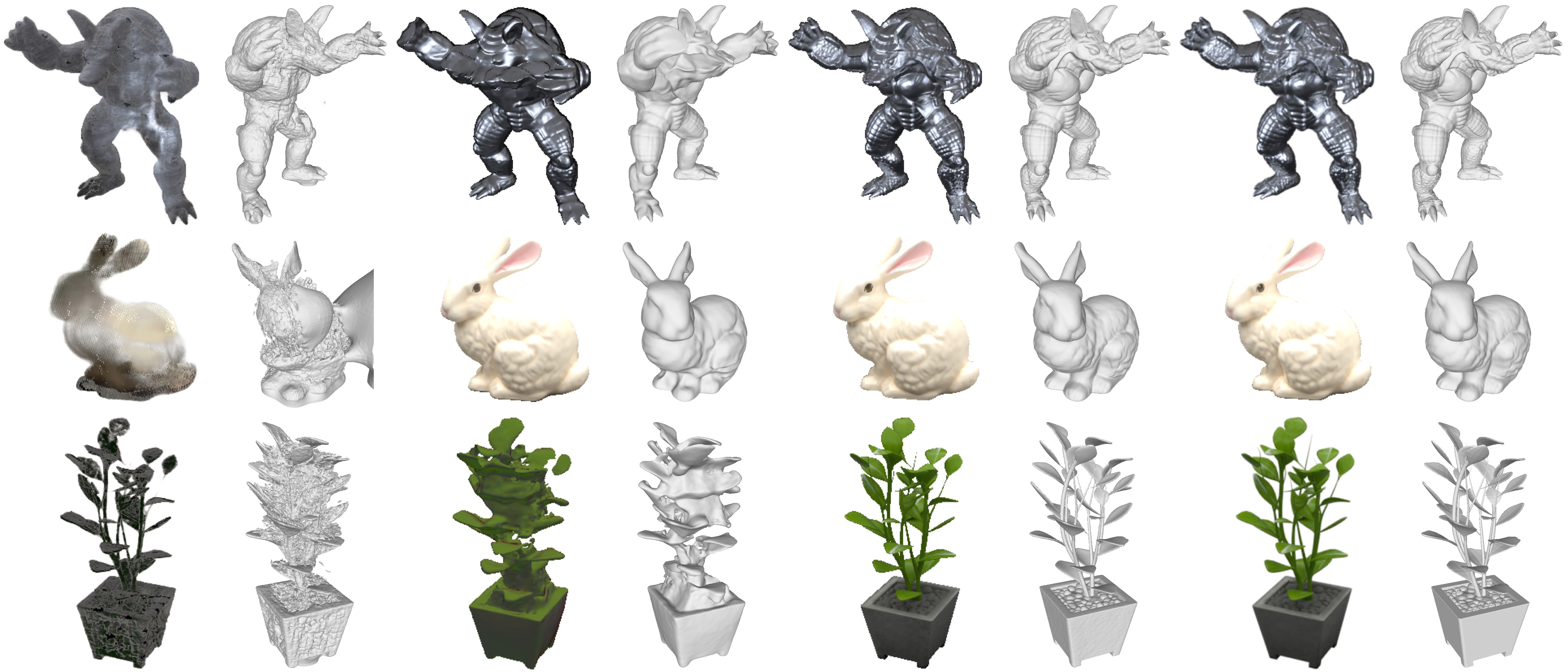}
    \begin{subfigure}[b]{0.24\textwidth}
    \caption{NeRD \cite{boss2021nerd}}
    \end{subfigure}
    \begin{subfigure}[b]{0.24\textwidth}
    \caption{PhySG \cite{zhang2021physg}}
    \end{subfigure}
    \begin{subfigure}[b]{0.24\textwidth}
    \caption{Ours}
    \end{subfigure}
    \begin{subfigure}[b]{0.24\textwidth}
    \caption{Ground truth}
    \end{subfigure}
    \caption{Visual comparison of novel rendering and surface geometry versus ground truth. NeRD uses a particle cloud geometry representation that results in bumpy surfaces. PhySG produces a hard, smooth surface but is unable to deal with complex geometries. Our method produces arguably better results than both baselines.}
    \label{fig:comparison}
\end{figure*}
Due to the diversity of imaging conditions, maintaining control variables (\eg camera pose and environment lighting) is difficult. We therefore compare with above methods on the synthetic data. All methods received HDR images, object segmentation masks, as well as camera poses. Specifically, we evaluate the commutative surface-to-surface distance between predicted and ground truth shapes on the visible regions.\footnote{See Appendix for the definition of surface-to-surface distance.} Additionally, we compute median and mean normal errors for all training view normal maps and depth maps. Results are listed in Table \ref{tab:quantitative_comparison}.

Each method is different in their own BRDF and illumination parameterization, therefore we compare the reflectance accuracy by rendering 20 novel-view and novel-lighting images under both flashlight and environment illumination. The results are listed in Table \ref{tab:novel_rendering}. Due to difference in setup, we swap the flashlight for a distant light source when evaluating \cite{boss2021nerd} and \cite{zhang2021physg}. A visual comparison is given in Figure \ref{fig:comparison}. Spatially varying BRDFs are better visualized with varying lighting conditions, and we refer the reader to our website for more visualizations.
\begin{table}
\small
\begin{threeparttable}
    \caption{Quantitative novel view and relighting results on synthetic scenes.}
    \setlength\tabcolsep{0.55em}
    \centering
\begin{tabular}{c*{6}{|c}}
\multirow{3}{*}{} & \multicolumn{2}{c|}{NeRD\footnotesize\cite{boss2021nerd}*} & \multicolumn{2}{c|}{PhySG\cite{zhang2021physg}} & \multicolumn{2}{c}{Ours}  \\
\cline{2-7}
                  & PSNR & SSIM  & PSNR & SSIM  & PSNR & SSIM   \\
                  \hline

 Armadillo &  21.27   &  0.811   &  33.86   &  0.957  & \textbf{40.39}  & \textbf{0.987}   \\
 Bunny &  15.69  &  0.799  &   19.33  &  0.953  &  \textbf{21.12}   &  \textbf{0.964}   \\
 LegoCar &   -  &  -   &   29.12  &  0.935  &   \textbf{38.40}  &  \textbf{0.986}    \\
 Plant &  20.66  &  0.819   &  15.15   &  0.924  &  \textbf{37.54}   & \textbf{0.965}  \\
\end{tabular}
\begin{tablenotes}[flushleft]
\small
     \item *Results of NeRD were obtained where ground truth images were used to solve for environment illumination maps.
\end{tablenotes}
    \label{tab:novel_rendering}
\end{threeparttable}
\vspace*{-0.1cm}
\end{table}
\subsection{Varying flashlight intensities}
The flashlight luminosity directly impacts how well the corresponding reflection model $\hat{\mathcal{L}}$ is supervised. When the flashlight suppresses ambient lighting, $\hat{\mathcal{L}}$ becomes dominant, effectively yielding darkroom images. On the other hand, if the flashlight is dimmed out, the flashlight model $\hat{\mathcal{L}}$, in particular reflectance predictions $\Theta$, lose all physical significance, and our method reverts to vanilla neural light field with NeuS \cite{wang2021neus} as backbone.\\
\begin{figure}
    \captionsetup[subfigure]{labelformat=empty}
    \centering
    \includegraphics[width=0.45\textwidth]{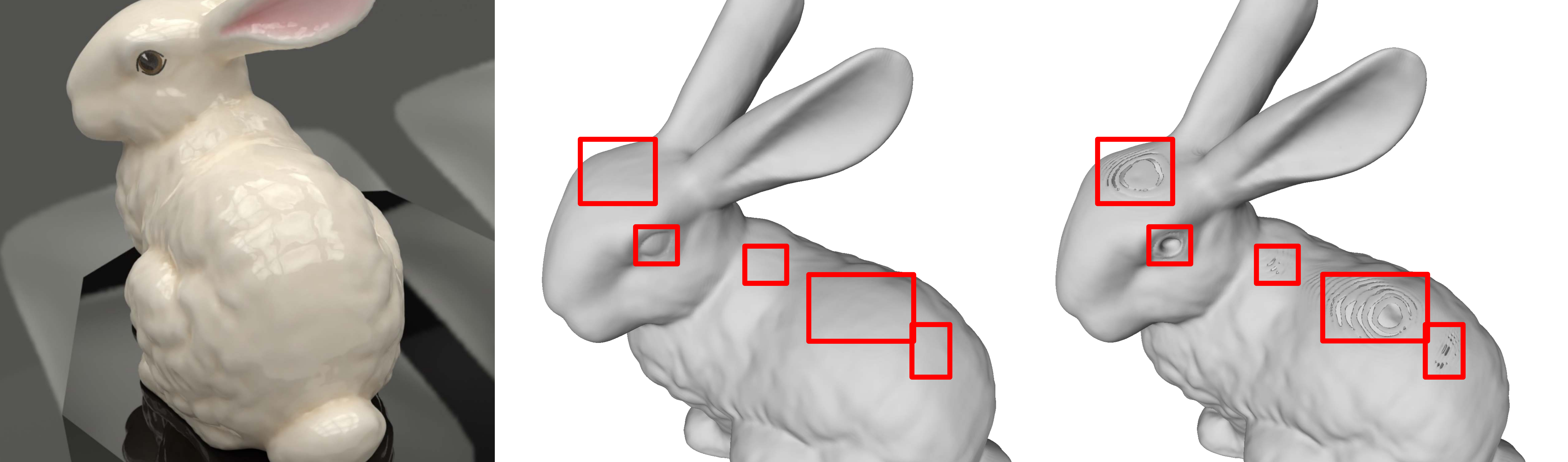}\\
    \begin{subfigure}[b]{0.15\textwidth}
    \caption{Input {\tiny (HDR tone mapped)}}
    \end{subfigure}
    \begin{subfigure}[b]{0.15\textwidth}
    \caption{Ours}
    \end{subfigure}
    \begin{subfigure}[b]{0.15\textwidth}
    \caption{NeuS \cite{wang2021neus}}
    \end{subfigure}
    \caption{Without flashlight, NeuS \cite{wang2021neus} creates holes on specular parts while our method is robust to specularities. Better viewed on screen zoom-in.}
    \label{fig:neus_comp}
\end{figure}
Figure \ref{fig:neus_comp} compares our result, produced with flashlight/non-flashlight images, to NeuS~\cite{wang2021neus} trained solely from non-flashlight images. NeuS can generate inconsistent geometry on specular regions due to inherent ambiguity between appearance and geometry. We were able to disambiguate this case by adding a physically-based flashlight reflection model and jointly solving BRDF and shape.\\
\begin{figure}[!tbp]
\begin{subfigure}[b]{0.09\textwidth}
         \centering
         \includegraphics[width=\textwidth]{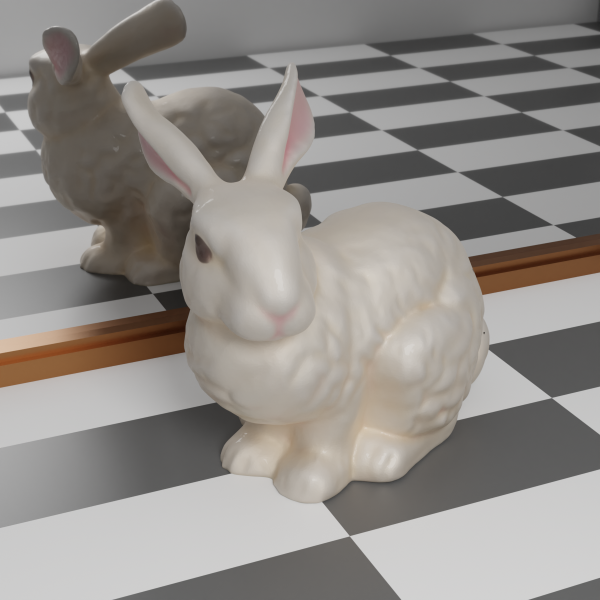}\\
         \includegraphics[width=\textwidth]{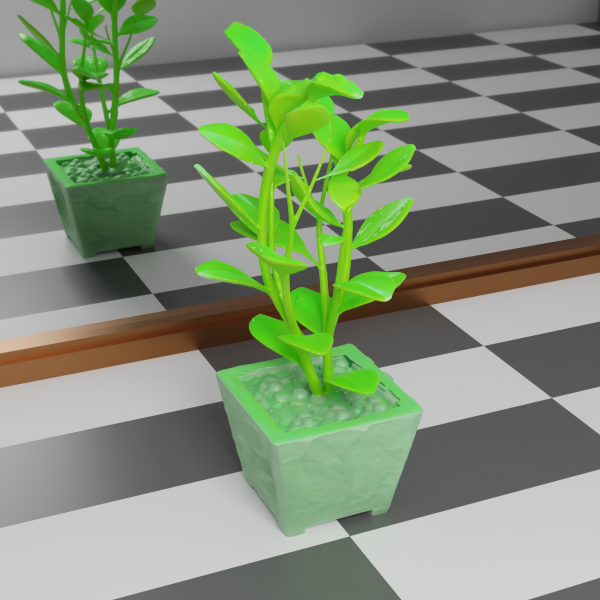}
         \caption{$ratio\ 0.2$}\label{fig:ablation_02}
     \end{subfigure}
\begin{subfigure}[b]{0.09\textwidth}
         \centering
         \includegraphics[width=\textwidth]{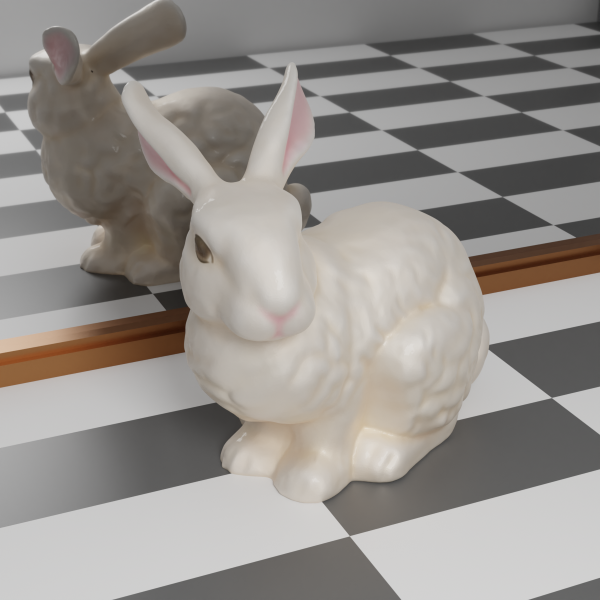}\\
         \includegraphics[width=\textwidth]{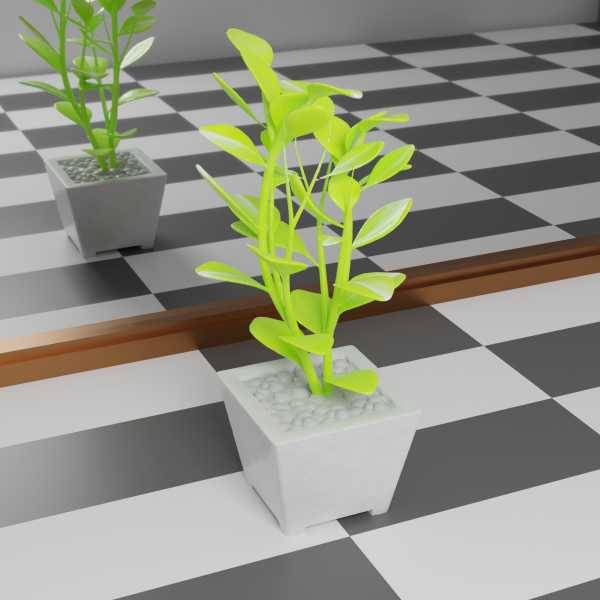}
         \caption{$ratio\ 0.4$}
     \end{subfigure}
\begin{subfigure}[b]{0.09\textwidth}
         \centering
         \includegraphics[width=\textwidth]{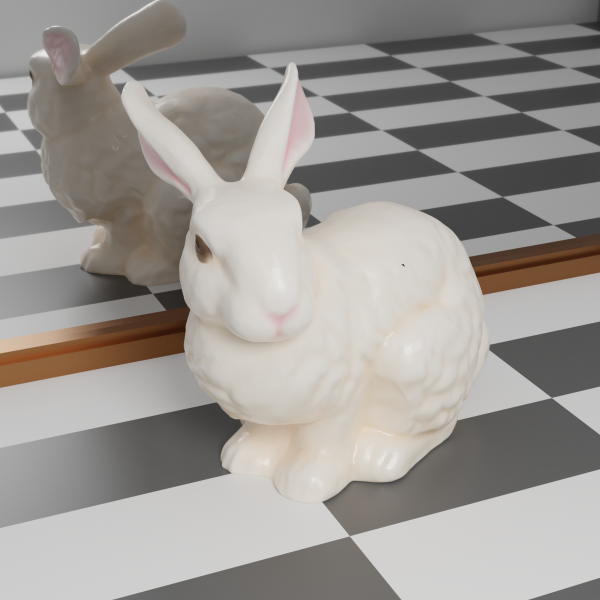}\\
         \includegraphics[width=\textwidth]{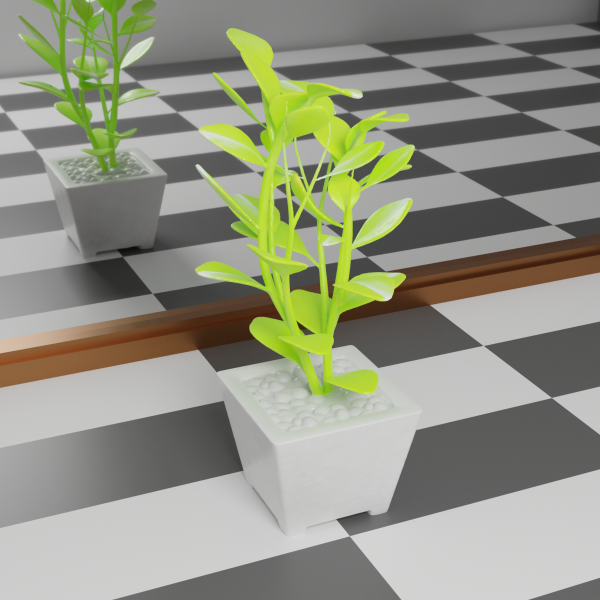}
         \caption{$ratio\ 0.6$}
     \end{subfigure}
\begin{subfigure}[b]{0.09\textwidth}
         \centering
         \includegraphics[width=\textwidth]{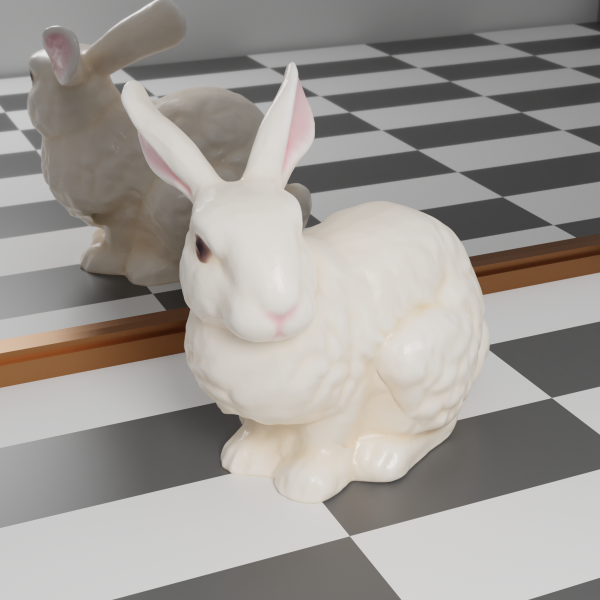}\\
         \includegraphics[width=\textwidth]{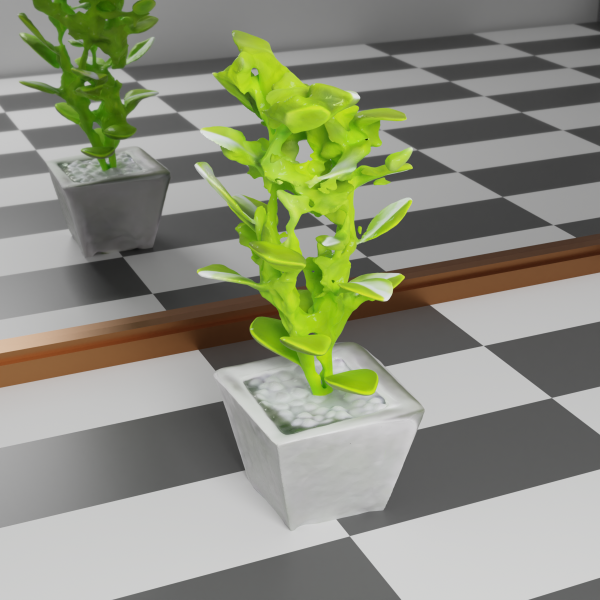}
         \caption{$ratio\ 0.8$}
     \end{subfigure}
\begin{subfigure}[b]{0.09\textwidth}
         \centering
         \includegraphics[width=\textwidth]{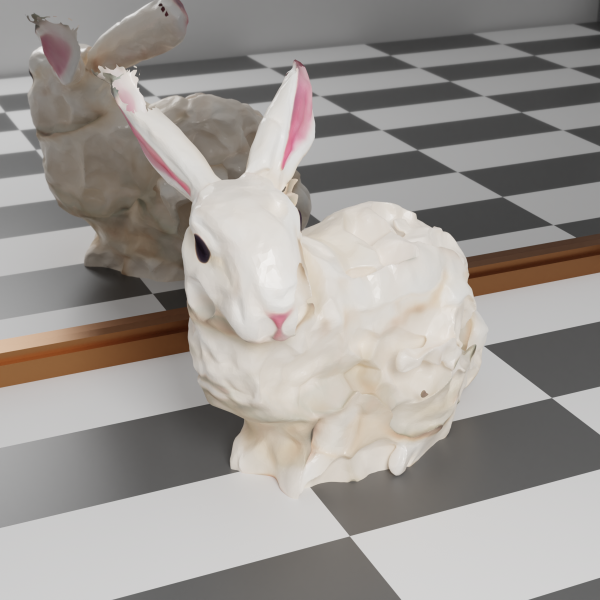}\\
         \includegraphics[width=\textwidth]{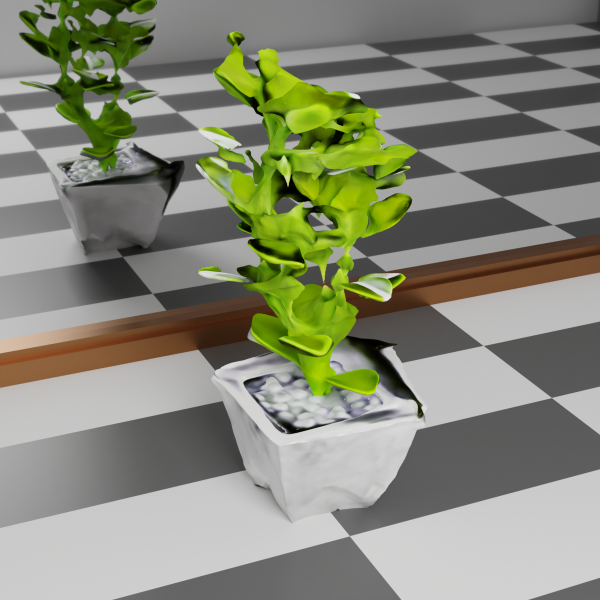}
         \caption{$ratio\ 1$}
         \label{fig:ablation_1}
     \end{subfigure}
     \caption{Reconstruction quality with varying flashlight intensities. Here $ratio\in[0.2,1]$ defines the strength of flashlight reflections over image intensities in the input images. That is, \ref{fig:ablation_02} shows reconstructed objects with the least bright flashlight, while \ref{fig:ablation_1} shows results in darkroom setting. Our method works best when the flashlight component makes up around half of total incoming lights.}\label{fig:ablation}
\end{figure}
Figure \ref{fig:ablation} illustrates the reconstructed objects at different ratios of flashlight component over total image intensities (\ie~$\text{ratio}=0$ means ambient-only, and $\text{ratio}=1$ means darkroom). When flashlight is dim, the reflectance parameters are insufficiently supervised and only geometry can be recovered. Interestingly, we see a considerable performance drop in geometry estimations when the flashlight intensity overwhelms ambient illuminations. We attribute this to three reasons: (a) previous work showed neural light fields harness empirically powerful priors for regularizing geometry \cite{yariv2020multiview}, however this prior is lost when ambient illumination becomes weak (b) the darkroom inverse rendering setup, although well-posed, is known to be highly non-convex \cite{nam2018practical,cheng2021multi} and hence sensitive to network initialization, and (c) in the darkroom setting, any non-zero output from ambient network becomes noise that disrupts optimization. Some potential solutions are to pre-train the intrinsic network on the object masks alone, and/or to involve additional geometry priors (\eg surface smoothness), and/or to scale down or disable ambient network. However, considering in real world the flashlight is unlikely to suppress environment illumination, we leave such improvement for future work.
\begin{figure}[!htb]
\centering
    Input \hspace{12mm} Geometry \hspace{16mm} Novel Renderings \hspace*{\fill}\\
    \hspace*{-0.6em}\includegraphics[width=0.5\textwidth]{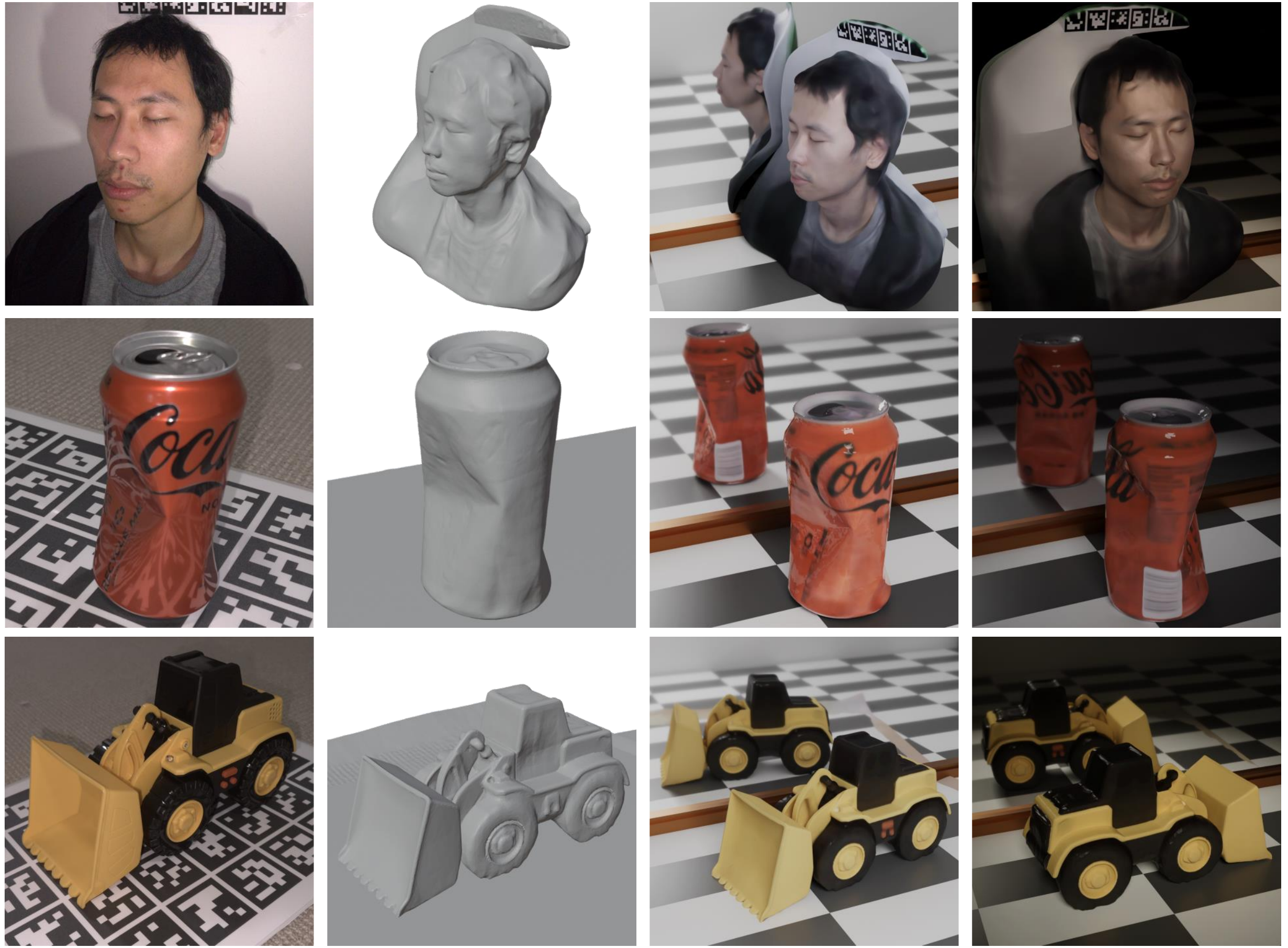}
    \caption{Visualization of reconstruction quality for three real indoor scenes.}
    \label{fig:realdata}
    \vspace*{-0.5cm}
\end{figure}
\subsection{Real world experiments}
Figure \ref{fig:realdata} illustrates our results on real world scenes. Unlike with the synthetic experiments before, here we do not provide object segmentation masks, but rely on the background NeRF to differentiate foreground objects from the background without manual intervention. The results show that partial scene inside the foreground regions of interest is well recovered with high fidelity.
\section{Conclusion}
In this paper we proposed a practical setup for obtaining scene geometry and reflectance. Compared to other in-the-wild inverse rendering methods, our approach does not explicitly decompose ambient reflections, but instead rely on the ambient and flashlight shaders for regularizing geometry and reflectance; we achieved arguably better reconstruction quality than competing state-or-arts. Two future directions to be explored are: (a) to extend our solution to large scenes with adjustable indoor lighting, and (b) to reconstruct translucent objects.

{\small
\bibliographystyle{ieee_fullname}
\bibliography{egbib}
}

\clearpage
\onecolumn
\appendix
\section*{Appendix}

\section{Principled BRDF formulation}
Our intrinsic network outputs a spacial distribution of texture elements (textels) $\Theta$ that parameterize the BRDF at that location. Since we are only interested in recovering opaque objects, we use a submodel of the original paper \cite{burley2012physically} without the refractive glass lobe. This parameterization can be expressed as:
\begin{align}
    \Theta = \{base\_color, roughness, clearcoat\_glossiness\nonumber\\, subsurface, metallic, dieletric, clearcoat\} \in {[0,1]}^{9}
\end{align}
where $base_color\in{[0,1]}^3$ is a 3D vector that defines the RGB base color of material, and all other terms are scalars in range $[0,1]$. Since the camera and light angle are always aligned in a co-located setup, hereafter we denote them by a single direction $\mathbf{h}$.

The model is a linear combination of diffuse and specular lobes, defined as follows:
\begin{align}
    \rho(\mathbf{n},\mathbf{h};\Theta) = &(1-metallic) \rho_\mathrm{diffuse}(\mathbf{n},\mathbf{h};\Theta) + \\\nonumber
    & metallic \rho_\mathrm{metallic}(\mathbf{n},\mathbf{h};\Theta) + \\ \nonumber
    & 0.08\times(1-metallic)dieletric \rho_\mathrm{dieletric}(\mathbf{n},\mathbf{h};\Theta) + \\ \nonumber
    & 0.25\times clearcoat \rho_\mathrm{clearcoat}(\mathbf{n},\mathbf{h};\Theta) \\ \nonumber
\end{align}
The diffuse component have two lobes: one for base diffuse and one for subsurface scattering
\begin{equation}
    \rho_\mathrm{diffuse} = (1-subsurface)\rho_\mathrm{base\_diffuse} + subsurface \rho_\mathrm{subsurface}
\end{equation}
where the two diffuse lobes are defined as
\begin{equation}
    \rho_\mathrm{base\_diffuse} = \frac{base\_color}{\pi} \Big(1+(2roughness-0.5)(1-\mathbf{n}^\top\mathbf{h})^5\Big)^2
\end{equation}
and $\rho_\mathrm{subsurface}$ is defined as
\begin{equation}
    \frac{1.25base\_color}{2\pi} \Big(\big(1+(roughness-1)(1-\mathbf{n}^\top\mathbf{h})^5\big)^2(\frac{1}{\mathbf{n}^\top\mathbf{h}}-1)+1\Big)
\end{equation}
respectively.

The metallic and dielectric lobes share the same GGX distribution, and both can be factorized into three terms:
\begin{equation}
    \rho_\mathrm{specular} = \frac{DGF}{4(\mathbf{n}^\top\mathbf{h})^2}
\end{equation}
where $D$ and $G$ are the microsurface distribution function and mask-shadowing term, respectively:
\begin{equation}
    D = \frac{roughness^4}{\pi \Big((roughness^4-1)(\mathbf{n}^\top\mathbf{h})^2 + 1\Big)^2}
\end{equation}
and
\begin{equation}
    G = \frac{2}{\sqrt{1 + \frac{roughness^4(1-(\mathbf{n}^\top\mathbf{h})^2)}{(\mathbf{n}^\top\mathbf{h})^2}} + 1}
\end{equation}
The only place where metallic and dielectric lobes differ is the Fresnel term $F$: the metallic lobe is chromatic while the dielectric lobe is not.
\begin{align}
    &F_\mathrm{metallic} = base\_color\\
    &F_\mathrm{dieletric} = 1
\end{align}

The clearcoat term can also be factorized similar to before
\begin{equation}
    \rho_\mathrm{clearcoat} = \frac{D_cG_cF_c}{4(\mathbf{n}^\top\mathbf{h})^2}
\end{equation}
where 
\begin{align}
    &D_c = \frac{roughness_c^2-1}{2\pi\log({roughness_c})(1+(roughness_c^2-1)(\mathbf{n}^\top\mathbf{h})^2)}\\
    &G_c = \frac{2}{\sqrt{1 + \frac{0.25^2(1-(\mathbf{n}^\top\mathbf{h})^2)}{(\mathbf{n}^\top\mathbf{h})^2}} + 1}\\
    &F_c = 0.2
\end{align}
Here the clearcoat roughness $roughness_c$ is confined within range $[0.001,0.1]$ and is linearly parameterized by $clearcoat\_glossiness$
\begin{equation}
    roughness_c = 0.1 - 0.099clearcoat\_glossiness.
\end{equation}
\section{Surface-to-surface distance for geometry evaluation}
We used a commutative mesh-to-mesh distance for evaluating the surface geometry produced from different method in Table 1. This distance metric is defined as follows:
\setlength{\abovedisplayskip}{3pt}
\setlength{\belowdisplayskip}{3pt}
\begin{equation}
    D(S_1, S_2) = \frac{1}{2} \mathop{\mathbb{E}}\big( d(\mathbf{x}_1,S_2) + d(\mathbf{x}_2, S_1)\big) \label{eq:distance}
\end{equation}
where $d$ is the point-to-manifold Euclidean distance, $S_1$ and $S_2$ are the visible surface regions, and $\mathbf{x}_1$ and $\mathbf{x}_2$ are two mutually independent random points uniformly sampled from $S_1$ and $S_2$ respectively. We compute the mean distance by repeatedly sampling $\mathbf{x}_1,\mathbf{x}_2$ until the standard variation of sampling mean is no greater than $10^{-5}$, or until an excessive number of one million pairs of points have been sampled. For the median distance, we replace the expectation operator $\mathop{\mathbb{E}}$ with the sample median. 

\end{document}